\def\year{2021}\relax
\documentclass[letterpaper]{article} 
\usepackage{aaai21}  
\usepackage{times}  
\usepackage{helvet} 
\usepackage{courier}  
\usepackage[hyphens]{url}  
\usepackage{graphicx} 
\usepackage{natbib}  
\usepackage{caption} 
\usepackage{amsmath,amssymb}
\usepackage{bbm} 
\usepackage{algorithm, algorithmic}
\usepackage{color}
\usepackage{multirow}
\usepackage{booktabs}       
\usepackage{amsfonts}       

\newcommand{\bqn}{\begin{eqnarray}}
\newcommand{\eqn}{\end{eqnarray}}
\newcommand{\beqarr}{\begin{eqnarray*}}
\newcommand{\eeqarr}{\end{eqnarray*}}
\newcommand{\bit}{\begin{itemize}}
\newcommand{\eit}{\end{itemize}}
\newcommand{\bs}{\boldsymbol}
\newcommand{\bx}{\mathbf{x}}

\newcommand{\bu}{\mathbf{u}}

\newcommand{\bsu}{\boldsymbol{u}}
\newcommand{\bst}{\boldsymbol{t}}
\newcommand{\umin}{\displaystyle\min}
\newcommand{\umax}{\displaystyle\max}

\newtheorem{definition}{Definition}
\newtheorem{theorem}{Theorem}

\newtheorem{example}{Example}
\DeclareMathOperator*{\argmin}{\arg\min}

\title{Kernel-convoluted Deep Neural Networks with Data Augmentation}
\author {
    Minjin Kim,\textsuperscript{\rm 1}
    Young-geun Kim, \textsuperscript{\rm 1}
    Dongha Kim, \textsuperscript{\rm 1}
    Yongdai Kim, \textsuperscript{\rm 2}
    Myunghee Cho Paik \textsuperscript{\rm 1}\\
}
\affiliations {
    \textsuperscript{\rm 1} Department of Statistics, Seoul National University \\
    \textsuperscript{\rm 2} School of Data Science, Seoul National University \\
    kmj1404@snu.ac.kr, cib009@snu.ac.kr, dongha0718@hanmail.net, ydkim0903@gmail.com, myungheechopaik@snu.ac.kr
}

\begin{document}

\maketitle

\begin{abstract}
The Mixup method \cite{zhang2018mixup}, which uses linearly interpolated data, has emerged as an effective data augmentation tool to improve generalization performance and the robustness to adversarial examples. The motivation  is to curtail undesirable oscillations by its implicit model constraint to behave linearly at in-between observed data points and promote smoothness. In this work, we formally investigate this premise, propose a way to explicitly impose smoothness constraints, and extend it to incorporate with implicit model constraints. First, we derive a new function class composed of kernel-convoluted models (KCM) where the smoothness constraint is directly imposed by locally averaging the original functions with a kernel function. Second, we propose to incorporate the Mixup method into KCM to expand the domains of smoothness. In both cases of KCM and the KCM adapted with the Mixup, we provide risk analysis, respectively, under some conditions for kernels. We show that the upper bound of the excess risk is not slower than that of the original function class. The upper bound of the KCM with the Mixup remains dominated by that of the KCM if the perturbation of the Mixup vanishes faster than \(O(n^{-1/2})\) where \(n\) is a sample size. Using CIFAR-10 and CIFAR-100 datasets, our experiments demonstrate that the KCM with the Mixup outperforms the Mixup method in terms of generalization and robustness to adversarial examples.
\end{abstract}

\section{Introduction}
Deep neural networks have brought an outstanding performance in various fields such as computer vision \cite{krizhevsky2012imagenet}, speech recognition \cite{graves2013speech}, and reinforcement learning \cite{silver2016mastering}. To train deep neural networks, we solve the empirical risk minimization (ERM) problem given data, but this solution could lead to a small training error, but a large test error, known as overfitting. This means that the deep neural networks could memorize a training sample, have poor generalization ability, and lack robustness against adversarial attacks. 

Among the many techniques for regularization to reduce overfitting, data augmentation has been widely used to improve generalization performance in machine learning. In particular, in image classification, various data augmentation methods such as horizontal reflections and rotations have been commonly applied \cite{krizhevsky2012imagenet, simard1998}. Recently, sample-mixed augmentation, called Mixup, has emerged as an effective tool \cite{zhang2018mixup}. The Mixup method generates virtual samples based on linear interpolations between random pairs of inputs and their corresponding labels. Trained deep models using the Mixup-generated samples have demonstrated superb performances in supervised learning \citep{zhang2018mixup,liang2018understanding}, unsupervised learning \cite{beckham2019adversarial,xie2019unsupervised}, and semi-supervised learning \cite{ berthelot2019mixmatch, verma2019interpolation}. The authors of the Mixup method \cite{zhang2018mixup} conjecture a poor generalization and lack of the robustness to adversarial examples may be due to unnecessary oscillations at in-between observed data points. The method is designed to encourage the model to behave linearly between training samples to curtail undesirable oscillations when predicting outside the training examples. This motivational claim is cogent, but has not been scrutinized.

In this paper, we formally investigate this premise, propose a way to explicitly impose constraints, and extend it to incorporate with implicit constraints. First, we propose a function class where constraints are explicitly imposed and provide a formal risk analysis. Second, we propose to incorporate the Mixup method in the proposed function class and provide corresponding risk analysis.

In the first part, our strategy starts by paying attention to the role of data augmentation in placing model constraints. For example, data augmentation techniques such as horizontal reflections and rotations encourage training to find invariant functions to corresponding transformations. Recent works have attempted to explicitly impose these constraints through careful construction of models \cite{cohen2016group, tai2019equivariant, van2018learning}. 
Using interpolated data instead of original samples, the Mixup method encourages the models to implicitly satisfy the linearity constraint.
In contrast to data augmentation for rotations or flipping, there has not been an effort to explicitly impose constraints by constructing models. We fill this gap and build models that explicitly bring desirable constraints of smoothness. We introduce a new function class composed of {\it kernel-convoluted models} (KCM), which is a derived function class given original functions. 
In the KCM, the smoothness constraint is directly imposed by locally averaging the original functions with a kernel function via convolution.
The Mixup constraints can be explicitly imposed in this  model when a proper data-dependent kernel function is used. To derive tangible theoretical results, we focus on data-independent kernels but allow various types of smoothing schemes with a different choice of kernels. We provide the upper bound of the excess risk in relation to the complexity of the original function class in the case of a linear function or a deep neural network. 

\begin{figure}[t]
    \centering
    \includegraphics[width=\columnwidth]{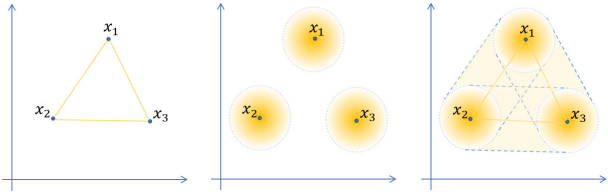}
    \caption{An illustration of regions where the Mixup method (left), KCM (middle), and KCM with the Mixup method (right) impose smoothness when there are three data points. 
    }
    \label{fig:illustration}
\end{figure}  

\begin{figure*}[t]
    \centering
    \includegraphics[width=17.5cm]{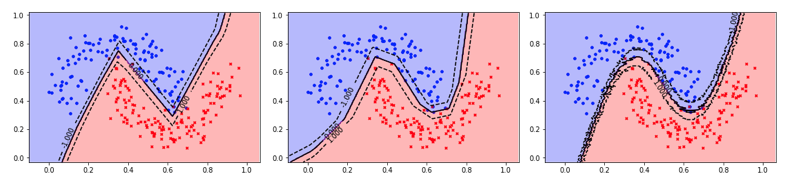}
    \caption{Visualization of the contours of classifiers learned with ERM (left), Mixup (middle), and KCM (right). Data points from classes -1 and 1 of the two-moon dataset are indicated by blue circles and red crosses, respectively. Decision boundaries are represented as solid lines and contours of levels -1 and 1 are represented by dashed lines. The best configuration of \(\alpha\) for Mixup and \(h\) for KCM is presented in Table 3 in the Supplementary material is presented. 
    }
    \label{fig:twomoon_plots}
\end{figure*}

In the second part of our paper, we incorporate the Mixup into the KCM and conduct a corresponding risk analysis. As opposed to the previous endeavors that replaced data augmentation by approaches explicitly modeling invariance to a certain transformation, we blend the two to expand the domain of smoothness.  While the KCM imposes smoothness over local regions around observed data defined by kernels as depicted by the yellow circles in the second subplot of Figure \ref{fig:illustration}, the Mixup method does over edges between two observed data as vertexes, as depicted by the yellow line in the first subplot of Figure \ref{fig:illustration}. Therefore, combining the two may expand the domains of smoothness, as depicted by the yellow triangular region in the third subplot of Figure \ref{fig:illustration}. Using the two-moon dataset \cite{scikit-learn}, we also show that the Mixup method promotes smoothness and the KCM renders the decision boundary visibly smoother than that of the Mixup method in Figure \ref{fig:twomoon_plots}.  For the minimizer of the risk of the KCM with the Mixup, we provide risk analysis and show that the upper bound of the excess risk can be expressed in terms of the complexity of the function class composed of KCM, and the smoothing parameter of the KCM, and the size of perturbation of the Mixup. Our main contributions are summarized as follows. 

\bit
    \item  We propose a new model called a kernel-convoluted model (KCM) where the smoothness constraint is directly imposed by locally averaging the original functions with a kernel function, and provide a risk analysis under some conditions for kernels.
    \item We propose to incorporate the Mixup method into the KCM to expand the domains of smoothness. We provide corresponding risk analysis and show that the upper bound of the excess risk is  \(O(n^{-1/2})\) when the original function is a deep neural network and  the perturbation order of the Mixup is faster than \(n^{-1/2}\) where \(n\) is a sample size (Theorem \ref{thm:excess_rb_mixup}). 
    \item Using CIFAR-10 and CIFAR-100 datasets, we demonstrate that the KCM with the Mixup outperforms the Mixup method in terms of generalization and robustness to adversarial examples. 
\eit

The rest of the paper is organized as follows: first, we review work related to the Mixup and efforts to build models that explicitly impose invariant constraints handled by data augmentation; then, we introduce a new function where smoothness is explicitly imposed and show its excess risk bound. Afterward, we propose to incorporate the Mixup into the proposed model and show the corresponding risk analysis; finally, we present experimental results and outline conclusions. We also provide some theoretical results, assumptions, and proofs in the Supplementary material.

\section{Related works and preliminaries}
\label{sec:related_works}

Since the Mixup has shown its effectiveness, many researchers have proposed variations and modifications. One kind of modification is to interpolate in the hidden space representations \citep{devries2017dataset, guo2019mixup, verma2019manifold}. Another variation is to choose more than a pair \cite{guo2019mixup}. \citet{tokozume2018between} consider alternative interpolation schemes involving labels. Moreover, \citet{liang2018understanding} use the spatial information to generate a new synthetic sample by stitching the space domain of two images with different proportions of the area of a synthetic image. 

There have been efforts to formalize data augmentation. One approach is to establish a theoretical framework for understanding data augmentation. For example, \cite{dao2019kernel} provide a general model of augmentation as a Markov process in which augmentation is performed via a random sequence of transformations. Other attempts include seeking an alternative yet formal way to replace the role of data augmentation. For example, one of the roles of data augmentation is to impose constraints of invariance on certain transformations. Some authors have proposed to formally construct a model to enforce desired transformation invariance constraints. In some images, rotation, flipping, and rescaling do not change labels. Such transformation invariances of a domain are often encouraged by data augmentation. Group invariance convolutional neural networks \citep{cohen2016group,dieleman2016exploiting,marcos2017rotation,worrall2017harmonic} enforces rotation or translation invariance. Equivariant transformer models \cite{tai2019equivariant} offer flexible types of invariance with the use of specially-derived canonical coordinate systems. \citet{van2018learning} demonstrate the explicit imposition of model constraint for the Gaussian process using the marginal likelihood criterion. In their works, the models are constructed in such a way that the predictions are invariant to transformations. Furthermore, \citet{van2018learning} argue the advantage of placing constraints directly through the model because the objective function correctly reflects the reduced complexity of constrained models. Recently, \citet{wu2020generalization} analyze the effect of the label-preserving transformations via the ridge estimator in an over-parametrized linear regression.
  
As for the Mixup in the classification problem, few studies have constructed a model to enforce desired constraints. This paper fills this gap. We present the models with explicit constraints that the Mixup attempts to attain and propose a class of models that achieves flexible types of smoothness.

\section{Kernel-convoluted neural networks and its excess risk bound}
\label{sec:KCM}

\subsection{Problem setup}

Before we describe the Mixup and study its constraints, we introduce the setup and notations.
We restrict our attention to a binary classification problem although the proposed method can be easily extended to a multi-class classification.
Let \(\mathcal{X}\subseteq \mathbb{R}^d\) be an input space and \(\mathcal{Y}\) be an output space. Assume that \(\mathcal{Y}=\{-1,1\}\). For a given real-valued function \(f\), we denote the classifier by \(C^*(\bx;f)\) where \(C^*(\bx;f)={\rm sign}\big(f(\bx)\big)\) for \(\bx \in \mathcal{X}\).
 
Other notations are as follows. For a real-valued function \(g\), we denote \(L_g\) by a Lipschitz constant of \(g\). The \(p\)-norm of \(\bx\) is defined by \(\|\bx\|_p=\big(|x_1|^p+\dots+|x_d|^p\big)^{1/p}\) where \(\bx=(x_1,\cdots,x_d)^T\) and \(p\ge 1\). For a matrix \(\bs A\in \mathbb{R}^{d\times m}\), we denote the spectral norm of \(\bs A\) by \(\|\bs A\|_2\).  
Given two positive real values \(a,b\), we write \(a\lesssim b\) if \(a\le cb\) for some generic constant \(c>0\). For two sequences \(\{a_n\}\) and \(\{b_n\}\), we write \(a_n=O(b_n)\) if there exists a positive real number \(M\) and a natural number \(n_0\) such that \(|a_n|\le M b_n{\rm { ~for ~all~ }}n \ge n_0\). 

To elicit implicit model constraints of the Mixup method, we briefly describe its steps \cite{zhang2018mixup}. Let the training data be \(S=\{(\bx_i,y_i)\}_{i\in [n]}\) drawn from unknown distribution \(\mathbb{P}_{\rm data}\).   For a randomly selected pair from the training data \(S\),  \((\bx_i,y_i)\) and \((\bx_j,y_j)\), consider a linear interpolation, 
\(\tilde{\bx}_i=(1-\lambda) \bx_i + \lambda \bx_{j}\), and  
\(\tilde{y}_i=(1-\lambda) y_i + \lambda y_j\),
where \(\lambda \in [0,1]\) is the interpolation weight. The Mixup method uses \(\{(\tilde{\bx}_i,\tilde{y}_i)\}_{i\in [n]}\) instead of originally observed data. 
By plugging in \((\tilde{\bx}_i,\tilde{y}_i)\), the trained model is forced to satisfy 
\(f(\tilde{\bx}_i)\approx (1-\lambda)f(\bx_i)+\lambda f(\bx_j)\). 
Thus, the Mixup method implicitly imposes the linear constraint, which induces  smoothing out the model to increase generalization power and robustness to adversarial examples.

\subsection{Kernel-convoluted models (KCM)}
\label{subsec:KCM_def}
In this section, we define kernel-convoluted models (KCM). Let $\mathcal{F}=\{f\mid f:\mathcal{X}\to \mathbb{R},~f~{\rm is~a~measurable~function}\}$.

\begin{definition} [Kernel-convoluted models]
Assume that \(\mathcal{X}\) is convex. For a real-valued function \(f\) and a measure defined on Borel \(\sigma\)-algebra \(\Sigma\) of subsets of \(\mathcal{X}\), \(K^*\), we call \(f^{K^*}\) kernel-convoluted models where 
\bqn
f^{K^*}\!(\bx)\!=\! K^*\!\ast f(\bx)\!=\!\!\int \! f(\bx-\bu)dK^*(\bu)~{\rm for}~\bx\in \mathcal{X}. \label{eqn:sem_measure}
\eqn
\end{definition}

The resulting function, \(f^{K^*}\),  is a weighted average of \(f\) at different input locations. If \(K^*\) is absolute continuous with respect to the Lebesgue measure \(\mu\), we denote the corresponding density by \(K\). Then, 
(\ref{eqn:sem_measure}) reduces to 
\bqn
f^{K^*}\!\!(\bx)\!=\!\!\int \!\! f(\bx\!\!-\!\!\bsu)K(\bsu) d\bsu\!=\!\!\int \! \!K(\bx\!\!-\!\!\bsu)f(\bsu)d\bsu. \label{eqn:sem_density}
\eqn

A convenient kernel to handle multivariate case is a product kernel defined by
\[K(\bu)=\prod_{j=1}^d k_h(u_j) =h^{-d}\prod_{j=1}^d k\Big(\frac{u_j}{h}\Big),\]
where \(k\) is a univariate kernel function, \(k_h(u)=h^{-1} k\big(u/h\big)\), and \(h>0\) is a bandwidth. When \(k\) is the Gaussian probability density function, the corresponding product kernel is called a \(d\)-dimensional Gaussian kernel.

With the above definition on kernel function
, \(h\) represents degree of heterogeneity of kernelizing models.  As \(h\) shrinks, the kernel-convoluted model converges to the original function:
\[\lim_{h\to 0} \int f(\bx-\bsu)K(\bsu)d\bsu=f(\bx).\]

\subsection{Generalization error bounds via the Rademacher complexity of KCM}
\label{subsec:KCM_geneerror}

We restrict our attention to  the case where \(K^{*}\) is absolutely continuous with respect to the Lebesgue measure \(\mu\) as (\ref{eqn:sem_density}) and suppress superscript $^{*}$ for brevit. 
We denote \(\mathcal{F}^{K}=\{f^{K} \mid f\in \mathcal{F}\}\) where $K$ is the corresponding density function. Assume that $K$ is the product kernel.
For a given real-valued kernel-convoluted function \(f^K\) defined on \(\mathcal{X}\), we consider a new classifier \(C(\bx;f^K)\) as \(C(\bx;f^K)={\rm sign}\big(f^K(\bx)\big)\) for \(\bx \in \mathcal{X}\). The classification risk of \(f^K\) is defined by 
\begin{align}
R(\mathbb{P}_{\rm data}, f^K)&=\mathbb{E}_{(\bx,y)\sim \mathbb{P}_{\rm data}}\left[I\big(C(\bx;f^K)\neq y\big)\right] \nonumber \\
&= \mathbb{E}_{(\bx,y)\sim \mathbb{P}_{\rm data}}\left[\ell\big(yf^K(\bx)\big)\right],\nonumber
\end{align}
where \(I(\cdot)\) is an indicator function and $\ell(z)=1$ if $z\leq 0$ and is 0 otherwise. While the 0-1 loss can be used for binary classification problem, directly minimizing the corresponding empirical risk is NP-hard due to the non-convexity of the $\ell$ \cite{hoffgen1995robust}. To resolve this issue, many authors have been considered a suitable surrogate loss that has advantageous  properties \citep{bartlett2006convexity, kim2018fast, mohri2012foundations, yin2019rademacher}. Denote a surrogate loss function \(\phi:\mathbb{R}\to [0,B]\). To avoid confusion, we add \(\phi\) to  a sub-index in the above definition of the risk and call it as \(\phi\)-risk: that is, for a given surrogate loss function \(\phi\), we define
\beqarr
R_\phi(\mathbb{P}_{\rm data}, f^K)  &=& \mathbb{E}_{(\bx,y)\sim \mathbb{P}_{\rm data}}\phi\big(yf^K(\bx)\big),\\
R_\phi(\mathbb{P}_{\rm data}, f^*_{\phi,\rm conv})&=&\inf_{f^K\in \mathcal{F}^K}R_\phi(\mathbb{P}_{\rm data}, f^K),\\
R_\phi(\mathbb{P}_{\rm data}, f_\phi^*)&=&\inf_{f\in \mathcal{F}}R_\phi(\mathbb{P}_{\rm data}, f).
\eeqarr

To estimate the \(\phi\)-risk we replace \(\mathbb{P}_{\rm data}\) by empirical distribution, \(\mathbb{P}_n\) and the surrogate empirical risk based on the training data \(S\) can be expressed as \(R_\phi(\mathbb{P}_n, f^K) = \frac{1}{n}\sum_{i=1}^n\phi\big(y_i f^K(\bx_i)\big).\) 

We use the empirical Rademacher complexity \cite{bartlett2002rademacher}, 
\[\hat{\mathcal{R}}_{S_z}(\mathcal{G})=\mathbb{E}_{\bs \epsilon}\Big[\sup_{g\in \mathcal{G}}\frac{1}{n}\sum_{i=1}^n\epsilon_ig(z_i)\Big],\]
to bound the generalization error
where \(\mathcal{G}\) is a family of functions mapping from \(\mathcal{Z}\) to \([a,b]\),  \(S_z=\{z_i\}_{i\in [n]}\) is a fixed sample of size \(n\), and \(\bs \epsilon=(\epsilon_1,\dots,\epsilon_n)^T\) and \(\epsilon_1,\dots,\epsilon_n\) are random variables with \(\mathbb{P}(\epsilon_i=1)=\mathbb{P}(\epsilon_i=-1)=1/2\) \cite{mohri2012foundations}. Let
\(\hat{f}_{\phi, \rm conv}=\argmin_{f^K\in \mathcal{F}^K}R_\phi(\mathbb{P}_n, f^K).\)  
Now, we present the excess risk of \(\hat{f}_{\phi, \rm conv}\) relative to \(R_\phi(\mathbb{P}_{\rm data},f^*_{\phi,\rm conv})\), and to \(R_\phi(\mathbb{P}_{\rm data},f^*_{\phi})\), respectively.  To avoid confusion, we call the former the excess KCM risk, and the latter, the excess risk. 
For the excess KCM risk, we can extend the results on the risks over \(\mathcal{F}\) via Rademacher complexity \cite{bartlett2002rademacher,mohri2012foundations}, and show that the upper bound of the excess KCM risk is bounded as long as the Rademacher complexity of \(\mathcal{F}^K\) is well-controlled. Details are in Corollary 1 in the Supplementary material. 

\begin{theorem}
\label{thm:excess_rb}
(Excess risk bound) Assume that the Lipschitz continuity on \(\phi\) and \(f\) stated in (A1) and (A2) in the Supplementary material as well as (A3) stated in below hold. For any \(\eta>0\), with probability at least \((1-\eta)\), we have
\begin{align*}
&R_\phi(\mathbb{P}_{\rm data}, \hat{f}_{\phi, \rm conv})-R_\phi(\mathbb{P}_{\rm data}, f^*_{\phi}) \\
&~~~~~ \le 4L_\phi \hat{\mathcal{R}}_{S}(\mathcal{F}^{K})+6B\sqrt{\frac{\log \frac{4}{\eta}}{2n}}+O(h).
\end{align*}
\end{theorem}

The first term of the upper bound is due to the Ledoux-Talagrand contraction inequality \cite{li2018tighter}, and  is a function of \(h\) through \(\mathcal{F}^{K}\).
Theorem \ref{thm:excess_rb} states that the excess risk bound depends on the Rademacher complexity and the degree of heterogeneity of kernerlizing models.
Therefore, a wide bandwidth can be a dominating term of the upper bound and leads to a slower convergence. The next section shows when \(f\) is a deep neural network, certain choices of kernels can keep the Ramemacher complexity of the new function class the same as that of the original function class asymptotically.

\subsection{The empirical Rademacher complexity of \(\mathcal{F}^K\)}
\label{subsec:KCM_empRademacher}

In this section, we study the upper bounds of \(\hat{\mathcal{R}}_{S}(\mathcal{F}^K)\) shown in Theorem \ref{thm:excess_rb} and its relationship with \(\hat{\mathcal{R}}_{S}(\mathcal{F})\). To delve into the effect of convolution, we first consider the case when \(f\) is a linear function with \(l_p\) norm constraints, and show that a symmetric univariate kernel function allows us to avoid polynomial dimension dependence even if \(p>1\) and to keep the same Rademacher complexity as that of the original class. Details are given in Theorem 2 in the the Supplementary material. 
Now, we characterize the KCM when the original function \(f\) is a deep neural network. For deep neural networks with \(L\) hidden layers, we denote a sequence of matrices by \(\mathcal{W}=\{\bs W_1,\dots,\bs W_{L+1}\}\), where \(\bs W_s\in \mathbb{R}^{d_s\times d_{s-1}}\) for \(s\in[L+1]\). Note that \(d_0=d\) and \(d_{L+1}=1\). Let 
\[f_{\mathcal{W}}(\bx)={\bs W}_{L+1} \rho\Big(\bs W_{L}\rho \big(\cdots \rho(\bs W_1\bx) \big)\Big),\] 
where \(\left[\rho(\bx)\right]_j=\rho(x_j)\) and \(\rho\) is an activation function. We use the rectifier linear unit (ReLU) activation function \(\rho(t)=\max\{0,t\}\) for \(t\in \mathbb{R}\). Let \(d_{\rm max}=\max\{d_0,d_1,\dots,d_{L+1}\}\). The loss function can be written as \(\phi\big(y f_{\mathcal{W}}(\bx)\big)\)
where \(\phi:\mathbb{R}\to [0,B]\) is \(L_{\phi}\)-Lipschitz. Define the function class for deep neural networks with spectrally-normalized constraints as follows:
\[
\mathcal{F}_{\rm DNN,\|.\|_{2}}=\{\bx \mapsto f_{\mathcal{W}}(\bx)\mid  \|\bs W_s\|_{2}\le r_s, s\in [L+1]\}. 
\]
We state the assumptions for the following theorem.
\bit
    \item [(A3)] For some constant \(c^1_K>0\), \(\int K_{\rm p}(\bsu)\|\bsu\|_2d\bsu<c^1_K< \infty\) where $K_{\rm p}(\bst)=\prod_{j=1}^d k(t_j)$.
    \item[(A4)] For any \(\bx \in \mathcal{X}\), \(\|\bx\|_2\le B_x\) for some constant \(B_x>0\).
\eit
Under (A4), \citet{li2018tighter} show that \(\hat{\mathcal{R}}_S\big(\mathcal{F}_{\rm DNN,\|.\|_{2}}\big)\lesssim G,\) 
where \[G=\frac{B_x\prod_{s=1}^{L+1}r_s }{\sqrt{n}}\sqrt{d_w\log \Bigg(\frac{(L+1)\sqrt{n}\umax_{1\le s\le L+1} \{r_s m_s\}}{\sqrt{d_w}\umin_{1\le s \le L+1} r_s}\Bigg)},\]
\(d_w=d_0\times d_1+\dots+d_{L}\times d_{L+1}\), \(m_s=\sqrt{{\rm rank}(\bs W_s)}\), and \({\rm rank}(\bs A)\) is the rank of a matrix \(\bs A\). Let \(\mathcal{F}^{K}_{\rm DNN, \|.\|_{2}}=\{\bx\mapsto K\ast f_{\mathcal{W}}(\bx)\mid f_{\mathcal{W}}\in \mathcal{F}_{\rm DNN,\|.\|_{2}} \}\). The following theorem shows the empirical Rademacher complexity of \(\mathcal{F}^K_{\rm DNN,\|.\|_{2}}\).

\begin{theorem}
Assume that 
\(K\) is a symmetric density function. Under (A3) and (A4), we have 
\[\hat{\mathcal{R}}_S\big(\mathcal{F}^{K}_{\rm DNN, \|.\|_{2}}\big)\lesssim \frac{B_*}{B_x}
G, \]
where \(B_*=3h c_K^1+B_x\).
\end{theorem}

We have analogous results  with the linear case in that  choosing symmetric  kernel functions,  the complexity of the new kernel-convoluted function class will be asymptotically equivalent to that of the original one.

\section{Kernel-convoluted models with Mixup}
\label{sec:KCM_mixup}

\subsection{Generalization error bound of KCM with Mixup}

In this section, we adapt the Mixup method to the KCM. A motivation of such adaptation is to broaden the domain of smoothness as depicted in Figure \ref{fig:illustration}. We conduct risk analysis of the proposed function class when the Mixup method is incorporated. The proposed estimator is  
\[\hat{f}_{\phi,{\rm prop}}=\argmin_{f^K\in \mathcal{F}^K}R_{\phi,{\rm prop}}(\mathbb{P}_n,f^K),\]
where \(R_{\phi,{\rm prop}}(\mathbb{P}_n,f^K)
=\frac{1}{n}\sum_{i=1}^{n} \phi\big( \tilde{y}_i f^K(\tilde{\bx}_{i})\big).\)
In this paper, we consider the case where the interpolation weight depends on \(n\), and is denoted by \(\lambda_n\). We show in Theorem 3 in the Supplementary material that the excess KCM risk with Mixup of \(\hat{f}_{\phi,{\rm prop}}\) defined by
\[R_\phi(\mathbb{P}_{\rm data},\hat{f}_{\phi,{\rm prop}})-R_\phi(\mathbb{P}_{\rm data},f^*_{\phi})\]
can be bounded by \(\hat{\mathcal{R}}_{S}(\mathcal{F}^{K})\), the term involving $\lambda_n$ and the bandwidth $h$. 

\begin{theorem}
\label{thm:excess_rb_mixup}
(Excess risk with Mixup)  
Let \(\mathcal{X}\) be a compact set. We assume that the Lipschitz constants, \(L_\phi\) and \(L_f\), are finite. Under (A1)-(A6) in the Supplementary material
for any \(\eta>0\), with probability at least \((1-\eta)\) we have:
\begin{align*}
&R_\phi(\mathbb{P}_{\rm data},\hat{f}_{\phi,{\rm prop}})-R_\phi(\mathbb{P}_{\rm data},f^*_{\phi})\nonumber \\
&~~~~\le 4L_\phi\hat{\mathcal{R}}_{S}(\mathcal{F}^{K})+6B\sqrt{\frac{\log \frac{4}{\eta}}{2n}}+O(\lambda_n)+O(h).
\end{align*}
\end{theorem}

Theorem \ref{thm:excess_rb_mixup} shows that the excess risk  with Mixup of \(\hat{f}_{\phi,{\rm prop}}\) can be bounded by the terms from the upper bound of the excess KCM risk with Mixup and the size of the bandwidth.  This implies that even with choosing a symmetric kernel, the dependence on \(h\) still remains.  When perturbation of the Mixup and smoothing range of KCM are smaller than \(O(n^{-1/2})\), the upper bound would be asymptotically equivalent to that of the original class. 

\begin{algorithm}[tb]
    \caption{Training Mixup with KCM}
    \label{alg:KCM}
	\begin{algorithmic}
   		\STATE {\bfseries Input:} Training data \(S=\{(\bx_i,y_i)\}_{i\in [n]}\), kernel density function \(K\), parameter of Beta distribution \(\alpha\), size of minibatch \(n_B\), sample size for Monte Carlo approximation \(N\)
   		\STATE {\bfseries Parameter:} Parameter \(\bs \theta\) in a given model \(f_{\bs \theta}\)
   		\STATE {\bfseries Output:} Updated parameter \(\hat{\bs \theta}\)	
   		\STATE Initialize parameters \(\bs \theta\) in the model \(f_{\bs \theta}\)   
   		\FOR{\(t=1\) {\bfseries to} \({\rm iter_{max}}\)}
   			\STATE \textbf{Randomization of} \(\lambda\): \(\lambda_j\sim {\rm Beta}(\alpha,\alpha)\) for \({j\in [n_B]}\) and put \(\lambda=\min\{(1-\lambda),\lambda\}\)
			\STATE  \textbf{Mixup}: for randomly selected pairs \((\bx_j,y_j)\) and \((\bx_{j'},y_{j'})\), construct 
			\(\tilde{S}_B=\{(\tilde{\bx}_j,\tilde{y}_j)\}_{j\in [n_B]}\),\\
 	        where \beqarr
 	        \tilde{\bx}_j&=&(1-\lambda_j) \bx_j +\lambda_j  \bx_{j'}\\
 	        \tilde{y}_j&=&(1-\lambda_j) y_j +\lambda_j  y_{j'}.
 	        \eeqarr
 	        \STATE \textbf{Realization of} \(\bu\): \(\bu_{i'}\sim K(\cdot)\) for \(i'\in [N]\)
 	            \FOR{\(j=1\) {\bfseries to} \(n_B\)}
 					\STATE \textbf{Averaging} \(\bar{f}_{\bs \theta}(\tilde{\bx}_j)=\frac{1}{N}\sum_{i'} f_{\bs \theta}(\tilde{\bx}_j-\bu_{i'})\)
 				\ENDFOR 
 				\STATE \textbf{Update \(\bs \theta\) by descending}:
 				\(1/n_B\sum_j \phi \big(\tilde{y}_j\bar{f}_{\bs \theta}(\tilde{\bx}_j)\big)\)
   		\ENDFOR
    \end{algorithmic}
\end{algorithm}

\subsection{Algorithm}

The algorithm for the proposed method is provided in Algorithm \ref{alg:KCM}.  
In implementing the KCM, the kernel-convoluted function can be approximated by Monte Carlo approximation,
\(f^K(\bx)\approx \hat{f}^K(\bx)= N^{-1}\sum_{i'=1}^{N}f(\bx-\bu_{i'})\) where \(\bu_{i'}\) are random samples from a kernel density function \(K\) for \(i'\in [N]\). We bring attention to the fact that the case of N being 1 gives an unbiased estimator of the integral. In our experiments on CIFAR-10 and CIFAR-100, we find that the proposed method with N=1 or 5 produces better test accuracies than the Mixup method as described in the next section. 

\section{Experiment}
\label{sec:experiment}

We conduct experiments using three datasets, the two-moon dataset \citep{scikit-learn}, CIFAR-10, and CIFAR-100 \cite{krizhevsky2009learning} for both binary classification and multi-class classification. The results for binary classification are summarized in Table 3 and 4 in Appendix 1 in the Supplementary material. 
For multi-class classification, we extend our binary classification setup by employing the cross-entropy loss function. Using the datasets, we compare (i) ERM, (ii) Mixup, (iii) KCM with various degrees of \(h\) and \(N\) 
and (iv) Mixup with KCM and compare the effect of implicit constraint, explicit constraint through the proposed KCM and combined constraint through the Mixup with KCM. We denote the Mixup method as `MIXUP' and the Mixup with KCM as `MIXUP+KCM'. We also conduct experiments for robustness to adversarial examples to show that MIXUP+KCM can significantly improve the robustness of neural networks in comparison to MIXUP. We use the \(d\)-dimensional Gaussian kernel. We provide implementation details and results in Appendix 1 in the Supplementary material. 

\subsection{CIFAR-10 and CIFAR-100}
\label{subsec: cifar10/100}
The CIFAR-10 dataset consists of 60000 RGB images in 10 classes, with 6000 images per class. The CIFAR-100 dataset is similar to CIFAR-10 except it has 100 classes containing 600 images each. Both datasets have 50000 training images and 10000 test images.

To make a direct comparison with the original Mixup using CIFAR-10/100, we adopt the experimental configuration in the Mixup paper \cite{zhang2018mixup} and use the author's official code \cite{zhang2018mixupgithub}. As for MIXUP+KCM, we add code for the local averaging part (the fourth line from the bottom in Algorithm \ref{alg:KCM}). 
We note that there are training and test phases, and the performance of the methods is measured by the median of test accuracies of the last 10 epochs as the Mixup paper considered. We use ResNet-34 which is one of the architectures from the official code.  

Table \ref{table: cifar10_100} reports the median test accuracies of the last 10 epochs with the best configuration for KCM and MIXUP+KCM. The detailed results are summarized in Table 5 in Supplementary material. 
In particular, in CIFAR-10, with an appropriate choice of \((h,N)\), KCM/MIXUP+KCM outperforms ERM/MIXUP. However, in CIFAR-100, consideration of KCM leads to better performance both ERM and MIXUP for all configurations.

\begin{table}[]
    \centering
    \begin{tabular}{ccc}
    \toprule
    Dataset & Learning rule & Test accuracy \\
    \cmidrule(l){1-3}
    \multirow{4}{*}{CIFAR-10} & ERM & 95.01  \\
    & KCM \((0.01,1)\) & 95.06  \\
    & MIXUP & 96.11  \\
    & MIXUP  + KCM \((0.01,5)\) & \textbf{96.39}  \\
    \cmidrule(l){1-3}
    \multirow{4}{*}{CIFAR-100} & ERM &  74.44  \\
    & KCM \((0.01,1)\) & 75.34 \\
    & MIXUP & 78.47 \\
    & MIXUP + KCM \((0.01,1)\) & \textbf{79.12} \\
    \bottomrule
    \end{tabular}
    \caption{(Test accuracy) The median test accuracies of the last 10 epochs.  
    For KCM, the configuration pair \((h,N)\) represents the combination of the bandwidth \(h\) and the sample size for Monte Carlo approximation \(N\). For MIXUP, we set $\alpha=1$. The full results are summarized in Table 5 in the Supplementary material.}
    \label{table: cifar10_100}
\end{table}

\subsection{Robustness to adversarial examples}

Neural networks trained using ERM are vulnerable to visually imperceptible but thoroughly chosen adversarial perturbations, which lead to deterioration of the performance of the model \cite{szegedy2014intriguing}. Data augmentation also has been commonly used to increase model robustness \citep{goodfellow2015explaining, zhang2018mixup, rajput2019does}. For example, \citet{rajput2019does} analyzes the effect of data augmentation via the lens of margin to provably improve robustness, especially the required size of the augmented data set for ensuring a positive margin.

To examine the robustness to adversarial examples and make a direct comparison to the Mixup method, we follow the experimental setting as the author employed in the original Mixup paper \cite{zhang2018mixup}. We use ResNet-34 model: two of them trained via ERM on CIFAR-10/100, the third trained using KCM, the fourth trained using the Mixup, and the fifth trained using  the Mixup with KCM. We administer white/black box attacks generated by FGSM and I-FGSM \cite{goodfellow2015explaining}. 
For every pixel, the maximum perturbation levels are 0.031 and 0.03 for CIFAR-10 and CIFAR-100, respectively. The number of iterations for I-FGSM is 10. The other setup is the same as that of the Mixup paper.   

The results are summarized in Table \ref{table: RtAE}. For all types of attacks, Top-1 test accuracies of KCM+MIXUP were higher than those of MIXUP with margins ranging from 0.24\% to 4.25\% in CIFAR-10, and with margins ranging from 0.09\% to 2.18\% in CIFAR-100.

\begin{table*}[t]
    \centering
    \begin{tabular}{ccccc}
    \toprule
    Dataset & Attacks & Learning rule & FGSM & I-FGSM \\
    \cmidrule{1-5}
    \multirow{4}{*}{CIFAR-10} & \multirow{2}{*}{White-box} & MIXUP \((\alpha=1.0)\) & 75.63 & 51.82\\
    & &  MIXUP \((\alpha=1.0)\) + KCM \((0.01,5)\) & \textbf{78.24} & \textbf{56.07} \\ \cmidrule{2-5}
    & \multirow{2}{*}{Black-box} & MIXUP \((\alpha=1.0)\) & 85.31 &	88.87\\
    & &  MIXUP \((\alpha=1.0)\) + KCM \((0.01,5)\) & \textbf{85.79} & \textbf{89.11} \\
    \cmidrule{1-5}
    \multirow{4}{*}{CIFAR-100} & \multirow{2}{*}{White-box} & MIXUP \((\alpha=1.0)\) & 38.79 &	22.99\\
    & &  MIXUP \((\alpha=1.0)\) + KCM \((0.01,1)\) & \textbf{39.65} & \textbf{25.17} \\ \cmidrule{2-5}
    & \multirow{2}{*}{Black-box} & MIXUP \((\alpha=1.0)\) & 63.62 &	67.97\\
    & &  MIXUP \((\alpha=1.0)\) + KCM \((0.01,1)\) & \textbf{63.79} & \textbf{68.06} \\
    \bottomrule
    \end{tabular}
    \caption{(Robustness) The test accuracies of each method based on Top-1 accuracy. For KCM, we choose the configuration pair \((h,N)\) that shows the best performance in Table \ref{table: cifar10_100}. }
    \label{table: RtAE}
\end{table*}

\section{Discussion and conclusions}
\label{sec:conclusion}
While the Mixup method reduces unnecessary oscillations or promotes smoothness by implicit linear constraints via data augmentation, we propose a new smoothed network, KCM. To expand the domain of smoothness, we adapt KCM to the Mixup method. We provide upper bounds of excess risk for the KCM and its adapted version with the Mixup. The results offer insights on how the degree of heterogeneity of kernelizing models and the size of perturbation in the Mixup play a role in risk analysis. The experimental results demonstrate that the Mixup method helps to smooth the decision boundary and improves accuracy, but the proposed method with explicit model constraints attains smoothness more effectively. Experiments on CIFAR-10 and CIFAR-100 show that the proposed method can improve performance and increase robustness to adversarial examples with proper parameters.

\section{Acknowledgments}
We are grateful to all the reviewers for their thoughtful comments and suggestions.

\begin{small}
\bibliography{KCM_ref}
\bibliographystyle{aaai21}
\end{small}

\end{document}